\documentclass[educ,sglanonrev]{informs-educ}

\usepackage{eqndefns-left}
\RequirePackage{tgtermes}
\RequirePackage{newtxtext}
\RequirePackage{newtxmath}
\RequirePackage{bm}
\RequirePackage{endnotes}
\usepackage{graphicx}
\usepackage{booktabs}
\usepackage{amsmath,amssymb}

\usepackage{algorithm}
\usepackage{algpseudocode}
\usepackage{tikz}

\usetikzlibrary{arrows.meta,positioning}

\usepackage[sort&compress]{natbib}
 \bibpunct[, ]{[}{]}{,}{n}{}{,}%
 \def\bibfont{\small}%
\EquationsNumberedThrough    

\ECRepeatTheorems  %

\MANUSCRIPTNO{EDUC-0001-2026.00}

\OneAndAHalfSpacedXII

\newcommand{\calN}{\mathcal{N}}

\newcommand{\E}{\mathbb{E}}
\newcommand{\R}{\mathbb{R}}
\newcommand{\W}[0]{\mathcal{W}_2}
\newcommand{\KL}[0]{\operatorname{KL}}
\newcommand{\TV}[0]{\operatorname{TV}}
\EquationsNumberedThrough    

\TheoremsNumberedThrough     

\MANUSCRIPTNO{}

\usepackage{url}
\usepackage{xcolor}

\newcommand{\rev}[1]{{\color{black}#1}}

\begin{document}

\RUNAUTHOR{Cheng and Zhu and Xie}
\RUNTITLE{Generative Models for Decision-Making under Distributional Shift}

\TITLE{Generative Models for Decision-Making under Distributional Shift}

\ARTICLEAUTHORS{%
\AUTHOR{Xiuyuan Cheng}
\AFF{Department of Mathematics, Duke University, Durham, NC, USA \\ \EMAIL{xiuyuan.cheng@duke.edu}}
\AUTHOR{Yunqin Zhu}
\AFF{H. Milton Stewart School of Industrial and Systems Engineering, Georgia Institute of Technology, Atlanta, GA, USA \\ \EMAIL{yzhu812@gatech.edu}}
\AUTHOR{Yao Xie}
\AFF{H. Milton Stewart School of Industrial and Systems Engineering, Georgia Institute of Technology, Atlanta, GA, USA \\ \EMAIL{yao.xie@isye.gatech.edu}}
}%

\ABSTRACT{%
Many data-driven decision problems are formulated using a nominal distribution estimated from historical data, while performance is ultimately determined by a deployment distribution that may be shifted, context-dependent, partially observed, or stress-induced. This tutorial presents modern generative models, particularly flow- and score-based methods, as mathematical tools for constructing decision-relevant distributions. From an operations research perspective, their primary value lies not in unconstrained sample synthesis but in representing and transforming distributions through transport maps, velocity fields, score fields, and guided stochastic dynamics. We present a unified framework based on pushforward maps, continuity, Fokker--Planck equations, Wasserstein geometry, and optimization in probability space. Within this framework, generative models can be used to learn nominal uncertainty, construct stressed or least-favorable distributions for robustness, and produce conditional or posterior distributions under side information and partial observation. We also highlight representative theoretical guarantees, including forward--reverse convergence for iterative flow models, first-order minimax analysis in transport-map space, and error-transfer bounds for posterior sampling with generative priors. The tutorial provides a principled introduction to using generative models for scenario generation, robust decision-making, uncertainty quantification, and related problems under distributional shift.
}%



\maketitle

\section{Introduction}
Modern data-driven decision problems are built around forecasts, yet decisions rarely depend only on a point prediction. They depend on an uncertainty distribution: its dependence structure, tail behavior, and the way it shifts across contexts, observations, or regimes. This is where generative models become relevant to operations research.

In standard predictive tasks in machine learning, one typically seeks to learn a mapping from covariates \(X\) to a target \(Y\). A familiar special case is the regression model
\(
Y = f(X) + \varepsilon,
\)
where $f$ is the unknown function and \(\varepsilon\) denotes a noise term, but more generally, the goal is to estimate a low-dimensional functional of the conditional law of \(Y \mid X\), such as a conditional mean, quantile, or class label. In this sense, much of conventional learning is \emph{discriminative}: its goal is accurate prediction of a specified target, rather than construction of a full data-generating mechanism. By contrast, a generative model aims to represent an entire probability distribution, either joint or conditional. A common formulation is
\[
Y = g_\phi(X,Z), \qquad Z \sim \pi_0,
\]
where \(\pi_0\) is a simple reference distribution. The learned object is therefore not merely a predictor, but a mechanism for constructing and sampling from a distribution.

This distinction is especially important in OR. Decisions under uncertainty are often sensitive not only to the center of a distribution, but also to dependence, tails, and regime variation. In many applications, historical data identify at best a nominal distribution \(P\), while the distribution relevant for planning, robustness, inference, or deployment is a different distribution \(Q\) that must be constructed, updated, or perturbed in a principled way. From this perspective, the value of generative models lies not only in their ability to reproduce observed data, but also in providing flexible tools for constructing the decision-relevant distributions that OR problems actually require.
\subsection{Why OR should care}

A large class of OR problems can be written as \(\min_{\theta\in\Theta}\E_{Y\sim Q}[\ell(\theta,Y)]\), where \(\theta\) is a decision, \(Y\) is a random input, and \(Q\) is the uncertainty distribution at deployment. In principle, the problem is clear: choose a feasible decision that performs well under \(Q\). In practice, however, the main challenge is often not only how to solve the optimization problem once \(Q\) is specified, but how to specify, estimate, or construct a distribution that meaningfully reflects the uncertainty the decision will actually face.

This gap between a nominal data-generating distribution and a decision-relevant deployment distribution appears throughout OR. In stochastic optimization, one needs scenario distributions that preserve dependence, tail behavior, and regime structure rather than merely matching marginal statistics. In robust planning, one needs adverse but plausible distributions that expose operational vulnerabilities without collapsing into arbitrary worst-case perturbations. In partially observed systems, one needs conditional or posterior distributions that update as new information arrives and support adaptive decisions over time. In all of these settings, the bottleneck is fundamentally distributional: decision quality depends on whether the model captures the right uncertainty law, not only on whether the downstream optimization problem is solved accurately.

Viewed this way, generative models enlarge the OR toolkit in three related ways. First, they support \emph{representation}: learning nominal uncertainty in settings where classical parametric models may fail to capture multimodality, nonlinear dependence, or structural constraints. Second, they support \emph{robustness}: generating stress scenarios or least-favorable plausible distributions through guided or adversarial modifications of a nominal law. Third, they support \emph{inference}: producing conditional or posterior distributions under side information, partial observation, or repeated updates. These three roles show that generative modeling is not peripheral to OR but directly connected to its core concerns of uncertainty representation, robust decision-making, and inference.

\subsection{Scope and roadmap}

This tutorial is about \emph{constructing decision-relevant distributions under distributional shift}. Its focus is not unconstrained sample synthesis, nor the use of generative models as generic tools for reproducing observed data. Instead, we study settings in which historical data provide a nominal distribution \(P\), while the distribution relevant for planning, robustness, inference, or deployment is a different distribution \(Q\) that must be constructed, updated, or perturbed in a principled way.

Our central viewpoint is that many such constructions can be understood as \emph{iterative algorithms in probability space}. At the intrinsic level, one may view distribution learning and distributional perturbation as optimization or evolution over probability distributions. At the constructive level, these updates are realized through transport maps, velocity fields, particle systems, or guided stochastic dynamics. This viewpoint provides the mathematical backbone of the tutorial, but the guiding question throughout is operational: \emph{what distribution does the decision need, and how should that distribution be constructed?}

The tutorial is organized around three visible OR roles for generative models: representation, robustness, and inference. We first formalize decision-making under distributional shift and introduce the mathematical tools needed to describe distribution evolution. We then present modern generative models as computational realizations of such distributions, develop representative theoretical guarantees, and turn to OR tasks under shift, including scenario generation, stress testing, conditional and posterior updating, and transport across regimes. We close with evaluation principles and a concluding discussion.

This chapter is not intended as a survey of generative modeling. Rather, it provides an OR-facing treatment of the subset most relevant to constructing uncertainty distributions for planning, robustness, and inference under distributional shift.

\subsection{Notation}

We work on \(\mathcal X=\R^d\). Let \(\mathcal P_2=\mathcal P_2(\R^d)\) denote the set of probability distributions on \(\R^d\) with finite second moment, and let \(\mathcal P_2^r:=\{P\in\mathcal P_2:P\ll dx\}\) denote those that admit densities. When \(P\in\mathcal P_2^r\) has density \(p\), we write \(dP(x)=p(x)\,dx\); when there is no ambiguity, we use \(p\) to denote both the density and the corresponding distribution.
For a distribution \(P\), let \(L^2(P)\) denote the space of measurable vector fields \(v:\R^d\to\R^d\) such that \(\|v\|_P:=\bigl(\int_{\R^d}\|v(x)\|^2\,dP(x)\bigr)^{1/2}<\infty\). For \(u,v\in L^2(P)\), define \(\langle u,v\rangle_P:=\int_{\R^d}u(x)^Tv(x)\,dP(x)\).
\rev{Intuitively, a pushforward distribution is the distribution obtained after applying a deterministic transformation to a random variable. For example, suppose \(U\sim P\) represents historical demand and \(T(u)=u+\delta\) represents a uniform upward demand shift with $\delta$. Then the shifted demand \(T(U)\) has distribution \(T_{\#}P\). Thus, \(T_{\#}P\) is not specified independently; it is generated by first sampling from \(P\) and then applying the map \(T\).}
For a measurable map \(T:\R^d\to\R^d\), the pushforward of \(P\) is denoted by \(T_\#P\), defined by \(T_\#P(A)=P(T^{-1}(A))\) for measurable sets \(A\). Equivalently, if \(Y\sim P\), then \(T(Y)\sim T_\#P\).

\section{Decision-making under distributional shift}

We now formalize the basic setup behind the tutorial: decisions are made under a deployment distribution \(Q\), while data typically identify only a nominal distribution \(P\). The central modeling question is how \(Q\) should be constructed from \(P\).

\subsection{Canonical setup: nominal \(P\), deployment \(Q\)}

We consider decision-making problems under uncertainty of the form
\begin{equation}
\min_{\theta\in \Theta}\; R_Q(\theta),
\qquad
R_Q(\theta) := \E_{Y\sim Q}[\ell(\theta,Y)],
\label{OR_model}
\end{equation}
where \(\theta\in\Theta\) is the decision variable, \(Y\) is a random input, \(\ell(\theta,Y)\) is the loss incurred by decision \(\theta\) under realization \(Y\), and \(Q\) is the probability distribution governing the uncertainty at deployment. Given observed samples \(Y_1,\dots,Y_n\), one typically replaces \(R_Q(\theta)\) by the empirical average
\[
\widehat R(\theta)=\frac1n\sum_{i=1}^n \ell(\theta,Y_i),
\]
yielding the sample average approximation (SAA) in operations research, equivalently, the empirical risk minimization (ERM) problem in statistics and machine learning. Depending on the application, \(\theta\) may represent an inventory level, reserve allocation, portfolio, or intervention policy, while \(Y\) may encode demand, renewable generation, market returns, latent states, or future trajectories. The quantity \(R_Q(\theta)\) is the deployment risk: it measures performance under the distribution that is operationally relevant when the decision is actually used.

\rev{A simple newsvendor example illustrates why the distinction between \(P\) and \(Q\) matters. Let \(\theta\) denote an order quantity and let \(Y\) denote random demand. With overage cost \(h\) and underage cost \(b\), the loss is
\[
\ell(\theta,Y)=h(\theta-Y)_+ + b(Y-\theta)_+ .
\]
If the deployment demand distribution is \(Q\), the optimal order quantity is the \(b/(b+h)\)-quantile of \(Q\). If one instead optimizes using the historical nominal distribution \(P\), then the decision may be systematically biased: when deployment demand shifts upward, the nominal solution under-orders and increases stockout cost; when demand shifts downward, it over-orders and increases excess-inventory cost. Thus, even in a classical OR model, the main issue is not only solving the optimization problem, but constructing the distribution \(Q\) under which the decision will actually be used.}

The central difficulty is that \(Q\) is rarely known directly. Historical data typically provide samples from a nominal distribution \(P\), reflecting past operating conditions, observed contexts, or a baseline data-generating regime. If one ignores the distinction between \(P\) and \(Q\), then one optimizes as though the historical distribution were also the relevant deployment distribution. In many applications, however, this is exactly where the model fails.

The relevant distribution may differ from the nominal one for several reasons: deployment may occur under a shifted regime; the decision-maker may wish to evaluate adverse but plausible scenarios rather than historical ones; new side information or partial observations may induce conditional or posterior updates; or transfer across populations, policies, or sensing environments may require correction of the baseline distribution. Thus, the modeling problem is not only to estimate a nominal distribution \(P\), but also to determine what distribution \(Q\) should govern the decision problem at hand.

This distinction between nominal and deployment distributions will serve as a recurring organizing principle throughout the tutorial. From this viewpoint, the role of modern generative models is not merely to reproduce samples from \(P\), but to provide constructive ways of representing, updating, perturbing, and sampling from decision-relevant distributions \(Q\). The central question is therefore: given historical information encoded by \(P\), what distribution does the decision actually need?

\rev{Table~\ref{tab:pq-shift-roadmap} summarizes several recurring sources of the \(P\)-to-\(Q\) gap that motivate the distribution-construction tools developed below.}

\begin{table}[t]
\centering
\footnotesize
\renewcommand{\arraystretch}{1.12}
\caption{Common reasons why the deployment distribution \(Q\) differs from the nominal distribution \(P\).}
\label{tab:pq-shift-roadmap}
\begin{tabular}{p{0.18\linewidth}p{0.4\linewidth}p{0.33\linewidth}}
\hline
\textbf{Type of shift} & \textbf{Why \(P\neq Q\)} & \textbf{Distribution needed} \\
\hline
Regime shift 
& Deployment occurs in a new region, season, market, or operating condition 
& Corrected deployment distribution, often \(Q=T_{\#}P\). \\

Stress shift 
& The task requires adverse but plausible scenarios beyond typical historical behavior 
& Stressed or least-favorable distribution. \\

Contextual shift 
& Uncertainty changes with observed covariates such as weather, demand signals, or network state 
& Conditional distribution \(Q=P(Y\mid X=x)\). \\

Partial observation 
& New noisy measurements reveal information about an unobserved state 
& Posterior distribution \(Q=P(\cdot\mid y)\). \\

Equilibrium shift 
& The distribution is generated by interacting agents' collective behavior 
& Endogenous equilibrium distribution. \\
\hline
\end{tabular}
\end{table}

\subsection{Wasserstein space and optimal transport}

We now introduce the Wasserstein-2 distance and the associated optimal transport map. For \(\mu,\nu\in\mathcal P_2\), let \(\Pi(\mu,\nu)\) denote the set of couplings of \(\mu\) and \(\nu\). The Wasserstein-2 distance is defined by
\begin{equation}\label{eq:ot}
W_2^2(\mu,\nu)
:=
\inf_{\pi\in\Pi(\mu,\nu)}
\int_{\R^d\times\R^d}\|x-y\|^2\,d\pi(x,y).
\end{equation}
When \(P,Q\in\mathcal P_2^r\) have densities \(p\) and \(q\), we also write \(W_2(p,q)\) for \(W_2(P,Q)\).

When \(\mu\in\mathcal P_2^r\), Brenier's theorem (see, e.g., \cite[Section 6.2.3]{ambrosio2005gradient}) implies that the infimum in \eqref{eq:ot} is attained by a deterministic map. More precisely, there exists a \(\mu\)-a.e.\ defined map \(T_\mu^\nu:\R^d\to\R^d\) such that the optimal coupling is given by \(\pi=({\rm Id},T_\mu^\nu)_\#\mu\). Equivalently, the Kantorovich formulation \eqref{eq:ot} agrees with the Monge formulation
\begin{equation}\label{eq:ot-monge}
W_2^2(\mu,\nu)
=
\inf_{T:\R^d\to\R^d,\; T_\#\mu=\nu}
\int_{\R^d}\|x-T(x)\|^2\,d\mu(x).
\end{equation}
The map \(T_\mu^\nu\) is called the optimal transport map from \(\mu\) to \(\nu\).

\subsection{Representing \(Q\) via transport}

The deployment distribution is often constructed from a nominal baseline through transport. Let \(P\) denote a baseline distribution on \(\R^d\), identified from historical data, and let \(U\sim P\). In this tutorial, we focus on transport-based constructions of decision-relevant distributions, namely distributions of the form
\[
Q=T_\#P,
\]
where $T$ is a mapping from $\R^d $ to $\R^d$.  Equivalently, if \(U\sim P\), then \(T(U)\sim Q\).

This viewpoint is useful when the deployment distribution should differ from the nominal one in a structured way while preserving aspects of the original geometry, such as dependence, temporal coherence, or spatial organization. It is also constructive: once a transport map \(T\) is specified or learned, the shifted distribution \(Q\) is immediately sampleable by drawing \(U\sim P\) and mapping it to \(T(U)\). The question is therefore not only which distributions are plausible, but how decision-relevant distributions can be generated from \(P\) through well-defined transformations. This viewpoint is somewhat different from the one commonly emphasized in the Wasserstein DRO literature, where the Wasserstein distance is typically introduced through the Kantorovich coupling formulation and uncertainty is modeled through an ambiguity set around a nominal distribution; see, for example, \citet{kuhn2019wasserstein}. Here, the emphasis is instead on constructive distribution shift through transport maps, which is the perspective most natural for generative modeling.

This perspective is broad enough to cover several types of distributional operations that arise in OR. A transport map may encode regime shift, adversarial perturbation, conditional updating, or posterior correction, depending on how \(T\) is constructed and constrained. In this sense, transport provides both a mathematical representation of distributional shift and a direct link to tractable algorithmic implementations through generative models.

The later sections develop this viewpoint further through dynamic formulations based on ODEs, SDEs, particle systems, and optimization in probability space.

\section{Mathematical background}

This section collects the basic mathematical facts used throughout the tutorial, \rev{and introduces three complementary views of distribution evolution. The continuity equation describes deterministic transport of probability mass and underlies flow models. The Fokker--Planck equation describes density evolution under stochastic dynamics and underlies diffusion and score-based models. The Benamou--Brenier formulation adds an optimization principle over density--velocity paths, linking transport to Wasserstein geometry. Together, these tools bridge the \(P\) to \(Q\) motivation to constructive methods for transporting, perturbing, and sampling decision-relevant distributions.}
\rev{Figure~\ref{fig:section3-roadmap} summarizes the relationships among the main mathematical tools introduced in this section.}
\begin{figure}[t]
\centering
\resizebox{0.95\linewidth}{!}{%
\begin{tikzpicture}[
  >=stealth,
  every node/.style={font=\small},
  box/.style={draw, rounded corners, align=center, inner sep=6pt, text width=3.55cm},
  arr/.style={->, thick}
]

\node[box] (ode) at (0,1.6) {ODE particles\\[2pt]
$\dot X_t=v_t(X_t)$};

\node[box] (ce) at (4.8,1.6) {Continuity equation\\[2pt]
$\partial_t\rho_t+\nabla\!\cdot(\rho_t v_t)=0$};

\node[box] (bb) at (9.6,1.6) {Benamou--Brenier\\[2pt]
optimize over $(\rho_t,v_t)$\\subject to continuity};

\node[box] (sde) at (0,-1.6) {SDE particles\\[2pt]
$dX_t=-\nabla V(X_t)\,dt+\sqrt{2}\,dW_t$};

\node[box] (fp) at (4.8,-1.6) {Fokker--Planck\\[2pt]
$\partial_t\rho_t=\nabla\!\cdot(\rho_t\nabla V+\nabla\rho_t)$};

\draw[arr] (ode.east) -- (ce.west);
\draw[arr] (sde.east) -- (fp.west);
\draw[arr] (fp.north) -- (ce.south);
\draw[arr] (ce.east) -- (bb.west);

\end{tikzpicture}%
}
\caption{\rev{Roadmap for Section~3. Particle dynamics induce density-evolution equations; the Fokker--Planck equation can be rewritten as a continuity equation; Benamou--Brenier adds an optimization principle over continuity-constrained paths.}}
\label{fig:section3-roadmap}
\end{figure}

\subsection{ODE and continuity equation}

We use \(x\in\R^d\) to denote a generic spatial variable; this should not be confused with the random variable \(X\) used earlier for context or predictors. An ordinary differential equation (ODE) describes deterministic evolution in \(\R^d\) through
\begin{equation}\label{eq:flownet}
\dot{x}(t)=v(x(t),t), \qquad t\in[0,T],
\end{equation}
where \(v:\R^d\times[0,T]\to\R^d\) is a time-dependent velocity field, and we also write \(v_t(\cdot)=v(\cdot,t)\). Under standard regularity assumptions on \(v\), such as Lipschitz continuity in the spatial variable, the ODE is well posed: for every initial value \(x(0)\), there exists a unique trajectory \(x(t)\), and the solution depends continuously on the initial value.

If the initial position is random, then the induced distribution evolves with time. Suppose $x(0) \sim P$, and $P$ has density $p$.
We denote by $\rho_t$ the density of $x(t)$. 
Then, under sufficient regularity, \(\rho_t\) satisfies the continuity equation
\begin{equation}\label{eq:liouville}
\partial_t \rho_t + \nabla\cdot(\rho_t v_t)=0,
\qquad \rho_0=p.
\end{equation}
Here \(\nabla\cdot u = \sum_{j=1}^d \partial_{x_j}u_j\) denotes the divergence of a vector field \(u=(u_1,\dots,u_d)\). Equation \eqref{eq:liouville} expresses conservation of mass: probability is transported by the velocity field \(v_t\), but neither created nor destroyed.

The continuity equation provides the basic link between particle dynamics and distribution evolution. In particular, if one can construct a velocity field \(v_t\) such that the solution \(\rho_T\) at time \(T\) is close to a target density \(q\), then the induced flow transports the initial distribution toward the target distribution. This is the basic mechanism underlying continuous-time flow-based generative models.

\subsection{SDE and Fokker--Planck equation}
\label{subsec:sde-fpe-review}

A stochastic differential equation (SDE) augments deterministic dynamics with random noise. A standard example is the Ornstein--Uhlenbeck (OU) process
\begin{equation}\label{eq:OU-SDE}
dX_t = -X_t\,dt + \sqrt{2}\,dW_t,
\end{equation}
where \(W_t\) is standard Brownian motion in \(\R^d\). More generally, we consider the diffusion
\begin{equation}\label{eq:diffusion-sde-2}
dX_t = -\nabla V(X_t)\,dt + \sqrt{2}\,dW_t,
\qquad X_0\sim P,
\end{equation}
where \(V:\R^d\to\R\) is a potential function. The OU
 process \eqref{eq:OU-SDE} corresponds to the special case \(V(x)=\|x\|^2/2\).

Let \(\rho_t\) denote the density of \(X_t\). Then \(\rho_t\) evolves according to the Fokker--Planck equation
\begin{equation}\label{eq:FPE}
\partial_t \rho_t = \nabla\cdot(\rho_t \nabla V + \nabla \rho_t).
\end{equation}
Equation \eqref{eq:FPE} is the stochastic analog of the continuity equation: it describes how probability mass evolves under both drift and diffusion.

The continuity equation \eqref{eq:liouville} and the Fokker--Planck equation \eqref{eq:FPE} are closely related. By comparing the two, we see that
\eqref{eq:FPE} can be rewritten as a continuity equation with a velocity field
\begin{equation}\label{eq:velocity_score}
v(x,t) = -\nabla V(x) - \nabla \log \rho_t(x).
\end{equation}
Thus, the same marginal evolution of distributions may be represented either through stochastic particle dynamics or through a deterministic probability flow. This relation was leveraged in the deterministic sampling method, namely DDIM \citep{song2021ddim} and probability flow ODE \citep{song2021score} in score-based diffusion models.
Specifically, once the score field $\nabla \log \rho_t$ is learned in the forward time diffusion process (training), the reverse time process (sampling) can be computed via integrating the ODE associated with the velocity field $v$ as in \eqref{eq:velocity_score}, and it induces the same marginal distributions of the particles as a stochastic (SDE-based) sampler.

\subsection{Dynamic optimal transport (Benamou--Brenier)}

Among all transport maps that push from a source distribution $P$ to a target $Q$, the {\it optimal transport} (OT) minimizes the transport cost under the Monge formulation of Wasserstein distance.

An alternative dynamic formulation of OT is given by the Benamou--Brenier representation \citep{benamou2000computational,villani2009optimal}. Instead of constructing transport through a single map, one may represent it through a time-dependent density \(\rho(x,t)\) and velocity field \(v(x,t)\):
\begin{eqnarray}\label{eq:Benamou-Brenier}
&&\inf_{\rho,v}\; \mathcal{T} := \int_0^1 \E_{x \sim \rho(\cdot,t)} \| v(x,t) \|^2 \, dt \\
&&\text{s.t.} \quad \partial_t \rho + \nabla \cdot (\rho v) = 0,
\qquad \rho(\cdot,0) = p,\quad \rho(\cdot,1) = q. \nonumber
\end{eqnarray}
Here \(\rho(\cdot,t)\) is the density at time \(t\), and the continuity equation enforces that \(\rho\) evolves under the velocity field \(v\). The quantity \(\mathcal T\) is the action or transport cost. \rev{For intuition, consider a one-dimensional distribution \(P\) and let \(Q\) be the same distribution shifted right by one unit. A simple transport path is \(\rho_t=(T_t)_{\#}P\) with \(T_t(x)=x+t\), under the constant velocity field \(v(x,t)\equiv 1\). The continuity equation describes the density moving smoothly from \(P\) to \(Q\), while the Benamou--Brenier formulation selects, among all such density--velocity paths, one with minimal total kinetic cost.}

Under suitable regularity conditions on \(p\) and \(q\), the minimum value of \(\mathcal T\) in \eqref{eq:Benamou-Brenier} equals the squared Wasserstein-2 distance between \(p\) and \(q\), and the minimizing velocity field can be interpreted as the optimal control of the transport problem.

This dynamic viewpoint will be important later: many generative models construct distributions by learning a time-dependent velocity field \(v_t\), thereby realizing transport through continuous-time evolution rather than through a single map.

\section{Algorithm basics}
\label{sec:algo-basics}

The previous sections introduced the main distribution-construction operations that arise in decision-making under distributional shift, namely transport, conditioning, and guidance. We now turn to the computational question: how can such distributional updates be realized in practice?

Modern generative models provide one answer. Rather than viewing them primarily as tools for unconstrained sample synthesis, we treat them here as \emph{computational realizations of probability distributions}. Depending on the model class, a distribution may be represented through an invertible map, a time-dependent velocity field, a score field, or stochastic dynamics. The goal of this section is not to survey architectures exhaustively, but to present the algorithmic primitives that will later connect generative modeling to optimization in probability space.

Generative models, from generative adversarial networks (GANs) \citep{GAN,WassersteinGAN,CGAN} and variational autoencoders (VAEs) \citep{kingma2013auto,VAE_review} to normalizing flows \citep{nflow_review}, have achieved broad empirical success and become central tools in modern machine learning. More recently, diffusion models \citep{song2019generative,ho2020denoising,song2021score} and closely related flow-based models \citep{lipman2023flow,albergo2023building,albergo2023stochastic,fan2022variational,xu2022jko} have attracted particular attention because of their high-quality generation in complex high-dimensional settings. Compared with score-based diffusion models, which are primarily designed for sampling, flow models have the additional advantage of direct likelihood evaluation, which is useful for statistical inference. At the same time, despite their empirical success, the theoretical understanding of these models remains incomplete. For the purposes of this tutorial, the most relevant point is that these methods provide concrete algorithmic mechanisms for constructing and transforming distributions.

\subsection{Score-based diffusion models}
\label{subsec:lit-sbdm}

Score-based diffusion models construct a generative model in two stages: a fixed forward diffusion that gradually perturbs data toward a simple reference distribution, and a learned reverse-time dynamics driven by an approximation of the score function. \rev{Intuitively, a score-based diffusion model learns how to reverse a controlled destruction of information. The forward diffusion step gradually adds noise to data until the distribution becomes close to a simple reference distribution, such as a standard Gaussian. This step is not meant to model the physical dynamics of the application; rather, it creates a sequence of intermediate noisy distributions for which a denoising direction can be learned. The reverse dynamics then start from the simple reference distribution and use the learned score field to move samples back toward the data distribution. Thus, forward diffusion turns data into noise in a prescribed way, while reverse dynamics turns noise back into structured samples.} 

\paragraph{Forward diffusion.}
A standard example is the variance-preserving DDPM process \citep{ho2020denoising,song2021score}, which generates a sequence \(\{X_n\}_{n=0}^N\) by
\begin{equation}\label{eq:VP-DDPM-1}
X_n = \sqrt{1-\beta_n}\,X_{n-1} + \sqrt{\beta_n}\,Z_{n-1},
\quad n=1,\cdots,N,
\end{equation}
where \(Z_n\sim\calN(0,I_d)\) are i.i.d.\ and \(X_0\sim P\) is drawn from the data distribution. As \(N\to\infty\), \eqref{eq:VP-DDPM-1} converges to the SDE
\begin{equation}\label{eq:VP-DDPM=SDE}
dX_t = -\tfrac12 \beta(t) X_t\,dt + \sqrt{\beta(t)}\,dW_t,
\quad t\in[0,T],
\end{equation}
where \(\beta(t)>0\) is a variance schedule and \(W_t\) is standard Brownian motion in \(\R^d\).

After a time reparametrization, \eqref{eq:VP-DDPM=SDE} reduces to the Ornstein--Uhlenbeck process \eqref{eq:OU-SDE}. Let \(\rho_t\) denote the density of \(X_t\). Then \(\rho_t\) evolves according to the Fokker--Planck equation \eqref{eq:FPE}.

\paragraph{Score learning and reverse dynamics.}
Diffusion models learn the score function
$
s_t(x):=\nabla \log \rho_t(x)
$
by score matching \citep{hyvarinen2005estimation,vincent2011connection}, typically through the mean-squared objective
\[
\int_0^T \int \|\hat s_t(x)-s_t(x)\|^2 \rho_t(x)\,dx\,dt.
\]
This objective scales well in high dimensions and is one of the main computational advantages of diffusion models.

Once the score model \(\hat s_t\) is learned, sampling is performed through a reverse-time dynamics starting from a simple reference distribution, typically \(\tilde X_T \sim \calN(0,I_d)\), with the goal that the terminal distribution of \(\tilde X_0\) is close to \(P\). Besides the reverse-time SDE, one may also use the deterministic probability flow ODE \citep{song2021score}. 
Its validity follows from the fact that the Fokker--Planck equation \eqref{eq:FPE} can be rewritten as the continuity equation \eqref{eq:liouville} with velocity field \eqref{eq:velocity_score}. Under suitable regularity, the stochastic diffusion and the deterministic probability flow induce the same family of marginal distributions. This relation underlies the close connection between diffusion models and flow-based generative models.

\subsection{Flow models}

Normalizing flows are a class of deep generative models that enable both efficient sampling and likelihood evaluation. Historically, flow-based models appeared earlier than diffusion models in the modern generative-modeling literature. Broadly speaking, normalizing flows can be divided into two categories: discrete-time flows and continuous-time flows.

\paragraph{Discrete-time flow models.}
Discrete-time flow models are based on compositions of invertible maps, often implemented through residual or coupling architectures \citep{he2016deep}. A typical prototype is
\begin{equation}\label{eq:resnet}
x_l = x_{l-1} + f_l(x_{l-1}), \quad l = 1, \cdots, L,
\end{equation}
where \(f_l\) is the neural network map parameterized by the \(l\)-th residual block, and \(x_l\) is the output of the \(l\)-th block. As written, \eqref{eq:resnet} is not automatically invertible; invertibility must be enforced either by architecture design or by additional regularity conditions. When this is done, the overall model defines a transport map between distributions.

\paragraph{Continuous-time flows.}
Continuous-time flows, also called continuous normalizing flows (CNFs), are formulated under the neural ODE framework \citep{chen2018neural}. In this case, the state \(x(t)\) evolves according to the ODE \eqref{eq:flownet}, where the time-dependent vector field is parameterized by a neural network. The discrete-time update \eqref{eq:resnet} may be viewed as a forward Euler discretization of \eqref{eq:flownet} on a sequence of time points.

In both cases, the model defines a deterministic transport between a data distribution and a reference distribution, often chosen to be a standard Gaussian \(Q=\calN(0,I_d)\), hence the name ``normalizing.'' Taking the continuous-time formulation \eqref{eq:flownet}, let \(P\) be the data distribution with density \(p\), let \(X(0)\sim P\), and denote by \(p_t\) the density of \(X(t)\). Then \(p_t\) satisfies the continuity equation \eqref{eq:liouville} with initial condition \(p_0=p\). If one can construct a vector field \(v_t\) such that \(p_T\) is close to the reference density \(q\), then the reverse-time flow transports samples from \(Q\) to a distribution close to \(P\).

A key issue is invertibility. In the continuous-time setting, invertibility is induced naturally by the ODE flow under standard regularity conditions, since the dynamics can be solved forward and backward in time. In the discrete-time setting \eqref{eq:resnet}, invertibility must be enforced either through special layer constructions, such as NICE, Real NVP, and Glow \citep{dinh2014nice,dinh2016density,kingma2018glow}, or through regularization mechanisms such as spectral normalization \citep{iResnet} or transport-cost regularization \citep{onken2021otflow,makkuva2020optimal,xu2022invertible}.

A notable advantage of flow models is that they admit likelihood evaluation. Although these computations may become challenging in high dimensions, the ability to evaluate log-likelihood is fundamentally useful, especially for maximum-likelihood training and statistical inference. A related idea also appears in diffusion models through the deterministic reverse dynamics given by the probability flow ODE \citep{song2021score}; once a diffusion model is trained, the corresponding likelihood can be evaluated through that representation as well.

\paragraph{Flow matching.}
Flow matching (FM) is a class of continuous-time flow models that avoids the simulation-dependent likelihood training used in continuous normalizing flows \citep{grathwohl2018ffjord}. Instead of optimizing a log-likelihood objective involving the divergence \(\nabla\cdot v_\theta\) along ODE trajectories, FM trains a neural velocity field by a simulation-free \(L^2\)-type matching loss. This greatly reduces computational cost in high dimensions and has made FM a competitive alternative to both likelihood-based flows and diffusion models \citep{lipman2023flow,albergo2023building,liu2022rectified,lipman2024flow}.

FM still adopts the continuous-time neural ODE framework on a time interval \([0,1]\). Let \(p\) denote the data distribution and \(q\) a reference distribution. One specifies an interpolation path
\begin{equation}\label{eq:interpolation}
\psi(t) := I_t(x_0,x_1), \qquad t\in[0,1],
\end{equation}
where \(x_0\sim p\) and \(x_1\sim q\). A common choice is linear interpolation between \(x_0\) and \(x_1\). The interpolation induces a family of intermediate distributions \(\rho_t\), namely the distributions of \(\psi(t)\).

The model trains a neural vector field \(\hat v(x,t)\) to match the instantaneous velocity of the interpolation path by minimizing
\begin{equation}\label{eq:fm_loss}
L(\hat v)
:=
\int_0^1 \E_{x_0,x_1}
\left\|
\hat v(\psi(t),t) - \frac{d}{dt}\psi(t)
\right\|^2 dt.
\end{equation}
Here we suppress the network parameterization for simplicity and view \(\hat v\) as an unconstrained vector field.

Although \eqref{eq:fm_loss} is defined through endpoint couplings and interpolation paths, its minimizer corresponds to a valid transport field. More precisely, call a velocity field \(v(x,t)\) \emph{valid} if the continuity equation with initial distribution \(p\) yields terminal distribution \(q\). Under suitable regularity assumptions on the interpolation family, there exists such a valid field \(v\) for which, up to an additive constant, the objective \eqref{eq:fm_loss} is equivalent to the weighted \(L^2\) loss
\[
\int_0^1 \int_{\R^d} \|\hat v(x,t)-v(x,t)\|^2 \rho_t(x)\,dx\,dt,
\]
where \(\rho_t\) is the marginal distribution induced by the interpolation path. Thus minimizing \eqref{eq:fm_loss} recovers a vector field whose continuity equation transports \(p\) to \(q\); see \citet{albergo2023building,lipman2023flow}.

An important feature of FM is that the prescribed path need not come from a diffusion process. Diffusion paths form one important class, in which case the induced distribution evolution coincides with that of a forward SDE. However, FM also allows non-diffusive probability paths, which can lead to simpler training and faster sampling. Related formulations include stochastic interpolants \citep{albergo2023building}, where the terminal distribution \(q\) may be arbitrary and accessible only through samples, and rectified flows \citep{liu2022rectified}. In this sense, FM provides a flexible transport-based framework for training continuous-time generative models without likelihood evaluation.

\paragraph{Consistency models (CMs).}
To improve the sampling efficiency of diffusion models, score-distillation approaches such as consistency models enable few-step or even one-step generation \citep{song2023consistency}. The basic idea is to learn a map \(f_\theta\) directly that sends a point \(x_t\) on the probability-flow ODE trajectory back to the corresponding clean sample \(x_0\), namely \(x_0=f_\theta(x_t,t)\), where \(x_0\sim p_{\rm data}\). Once trained, the model can generate samples in a single step by drawing \(x_1\sim \pi_0\) from the reference distribution and applying \(f_\theta\), or in a small number of steps to trade off speed and fidelity. In this sense, consistency models may be viewed as a distillation of diffusion dynamics into a direct transport map from the reference distribution back to the data distribution.

\subsection{Connection to map representation}

Once a velocity field is learned, the associated ODE induces a transport map from a reference distribution to a target distribution. More precisely, integrating \eqref{eq:flownet} from \(t=1\) to \(t=0\) defines a map \(T:\R^d\to\R^d\) such that  \(T_\# Q \approx P\). A similar interpretation applies to diffusion models. Although the forward particle dynamics are stochastic, the induced evolution of densities is deterministic and governed by the Fokker--Planck equation. When the score field \(s_t(x)=\nabla \log \rho_t(x)\) is known, the same marginal evolution can be represented by the probability flow ODE, which in turn defines an associated transport map.

Thus, flow-based and diffusion-based models may both be viewed as constructing maps from a simple reference distribution to a target distribution: the former directly through deterministic transport, and the latter either through stochastic dynamics or through an equivalent deterministic probability flow. Transport maps, velocity fields, and stochastic dynamics should therefore be understood as different but closely related representations of the same basic problem of transforming one distribution into another.

\rev{
Table~\ref{tab:generative-model-summary} summarizes the generative-model classes discussed in this section and the distributional objects they learn.
}

\begin{table}[h!]
\centering
\footnotesize
\renewcommand{\arraystretch}{1.12}
\caption{Generative-model classes discussed in Section~4.}
\label{tab:generative-model-summary}
\begin{tabular}{p{0.23\linewidth}p{0.35\linewidth}p{0.34\linewidth}}
\hline
\textbf{Model class} & \textbf{Mechanism} & \textbf{Object learned} \\
\hline
Diffusion models 
& Add noise forward, then learn reverse-time denoising dynamics 
& Score field or reverse dynamics mapping noise back to data. \\

Normalizing flows 
& Compose invertible maps or integrate an ODE 
& Transport map between a reference and target distribution. \\

Flow matching 
& Match velocities along interpolation paths 
& Velocity field whose continuity equation transports distributions. \\

Consistency models 
& Distill diffusion dynamics into few-step generation 
& Direct map from noisy states to clean samples. \\
\hline
\end{tabular}
\end{table}

\section{Extended algorithms for distribution shift}

The previous section focused on basic generative mechanisms for representing and transporting probability distributions. We now turn to more structured settings in which the target distribution is determined by the task rather than fixed in advance. These include transport between empirical distributions, robust generation, conditional and posterior updating, and equilibrium distributions arising from interacting-agent systems. We also describe a common particle-based implementation template that applies across several of these constructions.

\begin{table}[h!]
\centering
\footnotesize
\renewcommand{\arraystretch}{1.15}
\caption{Roadmap for the distribution-construction tasks in Section~5.}
\label{tab:section5-roadmap}
\begin{tabular}{p{0.26\linewidth}p{0.33\linewidth}p{0.33\linewidth}}
\hline
\textbf{Distribution-construction task} & \textbf{Example OR situation} & \textbf{Role of the deployment distribution \(Q\)} \\
\hline
Distribution-to-distribution transport & Transfer a demand, traffic, or outage model from one region, season, or operating regime to another & \(Q\) is a target-regime distribution obtained by transporting samples from a nominal source distribution \(P\). \\

Worst-case generation and robust construction & Stress testing a supply chain, portfolio, power grid, or service system under adverse but plausible conditions & \(Q\) is a stressed or least-favorable distribution constructed from \(P\) to expose vulnerabilities of a decision. \\

Conditional generation with observed context & Generate demand, renewable supply, travel time, or outage scenarios given weather, calendar, market, or network covariates & \(Q\) is a context-dependent conditional distribution, such as \(P(Y\mid X=x)\). \\

Posterior sampling and inverse problems & Update uncertainty about a latent state after partial observations, sensor readings, or noisy measurements & \(Q\) is a posterior distribution obtained by combining a prior model with an observation or likelihood model. \\

Mean-field game formulations & Model fleet repositioning, traffic routing, ride-sharing, or congestion systems with many interacting agents & \(Q\) is an endogenous equilibrium population distribution induced by agents' best responses. \\
\hline
\end{tabular}
\end{table}

The following subsections are not intended as a comprehensive survey. Rather, they present representative formulations showing how generative models can be used to construct decision-relevant distributions under different operational requirements.

\subsection{Distribution-to-distribution transport and domain-shift matching}

We begin with a basic generalization beyond standard generative modeling. In classical flow and diffusion models, one of the endpoint distributions is usually taken to be a simple reference distribution, such as a Gaussian. In many applications, however, the target distribution is itself nontrivial and need not admit a closed-form density; instead, it may only be accessible through samples. This leads to the problem of \emph{distribution-to-distribution transport}: given a source distribution \(P\) and a target distribution \(Q\) on \(\R^d\), construct a transport map that pushes \(P\) toward \(Q\). This setting is important in applications such as domain adaptation and transfer learning, where one seeks to adapt a model trained on a source domain to a target domain at lower cost by transporting source samples toward the target distribution \citep{courty2014domain,courty2017joint}. Related ideas have also been used in fairness, where transport is employed to adjust distributions across groups in a controlled manner \citep{silvia2020general}.

Suppose we are given two sample sets \(\{X_i\}_{i=1}^N\) and \(\{\tilde X_j\}_{j=1}^M\), drawn i.i.d.\ from \(P\) and \(Q\), respectively. As one example, one may use a flow-based model 
to learn a continuous invertible transport between \(P\) and \(Q\) from these two datasets. The model is based on the neural ODE \eqref{eq:flownet}, with velocity field \(v(x,t;\theta)\), which induces a time-dependent flow of distributions \(\rho_t\), with \(\rho_0=P\) and \(\rho_1\approx Q\). Equivalently, the ODE induces a terminal transport map \(T:\R^d\to\R^d\) such that \(T_{\#}P\approx Q\). From the optimal-transport viewpoint, this may be interpreted as learning a dynamic transport path between the two endpoint distributions. Since in practice both \(P\) and \(Q\) are observed only through samples, the endpoint constraints in the dynamic OT formulation \eqref{eq:Benamou-Brenier} are enforced only empirically, for example, through Kullback--Leibler (KL) divergence-based matching losses, with KL divergence defined in Section 6.1; see \citet{ruthotto2020machine,makkuva2020optimal,xu2023computing}. In this way, the usual generative-model setting involving a simple Gaussian reference distribution is extended to transport between an arbitrary pair of sample-defined distributions.

In the distribution-to-distribution setting, the target distribution \(Q\) is specified, at least through samples. We next consider a more implicit setting, in which the target distribution is not given in advance but is defined adversarially through a worst-case objective.

\subsection{Worst-case generation and robust distribution construction}

In many engineering and operational settings, the scenarios that matter most for reliability are not the typical ones but rare, high-impact events at the edge of the nominal data distribution. Examples include unusual road conditions in autonomous driving, extreme contingencies in power systems, and atypical but consequential cases in healthcare. Such events are often poorly represented in historical data, yet they are precisely the cases that drive stress testing, robustness evaluation, and risk-sensitive decision-making. This motivates the problem of \emph{worst-case generation}: rather than sampling from a nominal distribution, one seeks a perturbed distribution that emphasizes adverse but plausible outcomes.

A natural framework for this task is distributionally robust optimization (DRO), in which one evaluates performance against distributions lying in an ambiguity set around a reference distribution \(P\). Among the many DRO formulations, Wasserstein DRO is especially attractive because the Wasserstein metric encodes a geometry on the sample space and leads to uncertainty sets that reflect structured perturbations of the data-generating distribution. While much of the DRO literature focuses on worst-case values and robust decisions, our emphasis here is slightly different: we view the inner maximization as a mechanism for \emph{constructing} a worst-case distribution. This interpretation connects robust optimization to generative modeling.

 Wasserstein DRO provides a classical framework for such robust distribution construction; see, e.g., \citet{mohajerin2018data,blanchet2019quantifying,gao2023distributionally}. Let \(\ell(\theta,x)\) be a loss function, where \(\theta \in \R^p\) denotes the decision variable or model parameter and \(x \in \R^d\) the uncertain input. The Wasserstein DRO problem takes the form (see, e.g., \citep{xu2024flow,cheng2025worst,zhu2024distributionally,wen2026distributionally}),
\begin{eqnarray}\label{eq:wass-minimax-0}
&&\min_{\theta \in \R^p}\max_{Q \in \mathcal B_\delta}
\E_{x\sim Q}\,\ell(\theta,x),\\
&&\mathcal B_\delta := \{Q \in \mathcal P_2 : \W(Q,P)\le \delta\}, \nonumber
\end{eqnarray}
where \(P\) is the reference (nominal) distribution and \(\mathcal P_2\) denotes the set of probability measures on \(\R^d\) with finite second moment. A convenient penalized form is
\begin{equation}\label{eq:wass-minimax-1}
\min_{\theta \in \R^p}\max_{Q \in \mathcal P_2}
\left\{
\E_{x\sim Q}\,\ell(\theta,x)
-\frac{1}{2\lambda}\W^2(P,Q)
\right\},
\end{equation}
where \(\lambda>0\) controls the effective size of the perturbation. This formulation makes explicit the tradeoff between adversarial loss and transport cost: the worst-case distribution should increase the loss, but only through a geometrically plausible shift away from \(P\).

When the reference distribution \(P\) has a density, the inner maximization admits an equivalent transport-map formulation. Writing \(Q=T_{\#}P\) for a measurable map \(T:\R^d\to\R^d\), one obtains
\begin{eqnarray}
&&\max_{Q \in \mathcal P_2}
\left\{
\E_{x\sim Q}\,\ell(\theta,x)
-\frac{1}{2\lambda}\W^2(P,Q)
\right\}\\
&&=
\max_{T \in L^2(P)}
\E_{x\sim P}
\left[
\ell(\theta,T(x))
-\frac{1}{2\lambda}\|T(x)-x\|^2
\right],
\end{eqnarray}
where
\[
L^2(P):=\Bigl\{T:\R^d\to\R^d:\E_{x\sim P}\|T(x)\|^2<\infty\Bigr\}.
\]
Hence, the robust distribution-construction problem may be written as the minimax problem
\begin{eqnarray}\label{eq:wass-minimax-2}
&&\min_{\theta \in \R^p}\max_{T \in L^2(P)}
L(\theta,T),\\
&&L(\theta,T):=
\E_{x\sim P}
\left[
\ell(\theta,T(x))
-\frac{1}{2\lambda}\|T(x)-x\|^2
\right]. \nonumber
\end{eqnarray}
Here \(\lambda>0\) is a regularization parameter that controls the size of the adversarial perturbation: larger \(\lambda\) permits more aggressive shifts away from the reference distribution \(P\), while smaller \(\lambda\) keeps \(T(x)\) closer to \(x\). This representation is especially useful from a generative viewpoint: the worst-case distribution is no longer an abstract optimizer \(Q\), but the pushforward of the nominal distribution under an adversarial transport map \(T\). In this way, Wasserstein DRO can be interpreted as a problem of robust distribution construction, with the map \(T\) generating informative worst-case samples from the reference distribution.

\subsection{Conditional generation with observed context}

A different but related form of distribution construction arises when one seeks to generate \(Y\) given observed context \(X=x\), that is, to model the conditional distribution \(P(Y\mid X=x)\). This setting appears naturally in probabilistic prediction, contextual decision-making, and time-series forecasting. As in the unconditional case, the generative mechanism may be made context-dependent. In a flow-based formulation, one introduces a transport map \(T_0(\cdot;x)\) such that \(T_0(\cdot;x)_{\#}\pi_0\) approximates \(P(Y\mid X=x)\), where \(\pi_0\) is a simple reference distribution. In a continuous-time formulation, one instead uses a context-dependent velocity field \(v_t(\cdot;x)\), or, in the diffusion setting, a score field \(s_t(\cdot;x)\), so that the induced distribution evolution \(\rho_t(\cdot\mid x)\) connects a reference distribution to \(P(Y\mid X=x)\). Representative examples include conditional normalizing flows, conditional diffusion models, and conditional flow matching \citep{trippe2018conditional,ho2022classifier,generale2024conditional}. This should be distinguished from posterior updating in inverse problems and Bayesian inference: here one learns a predictive family of conditional distributions indexed by \(x\), whereas posterior conditioning updates a prior distribution through an observation model.

\subsection{Posterior sampling with generative priors and inverse problems}

Posterior sampling with generative priors has been studied in several forms, including invertible-model approaches and score-based or diffusion approaches to inverse problems; see, e.g., \citet{ardizzone2019analyzing,song2021score,song2022solving,chung2023diffusion}.

Let \(P\) be a prior distribution on \(\R^d\), let \(x\sim P\) denote the unknown state, and let \(y\) be an observation generated from a known likelihood \(p(y\mid x)\), assumed differentiable in \(x\). Conditioning updates the baseline distribution \(P\) to the posterior distribution \(P(\cdot\mid y)\), which reweights \(P\) according to the data-fidelity term induced by the observation model. A standard example is an inverse problem, where \(y=\mathcal A(x)+n\) with \(\mathcal A:\R^d\to\R^m\) a possibly nonlinear forward operator and \(n\) observation noise. Writing \(\mathcal L_y(x):=-\log p(y\mid x)\), the posterior density, when it exists, is proportional to \(e^{-\mathcal L_y(x)}p(x)\), where \(p\) denotes the density of \(P\).

When the prior distribution is represented by a pre-trained invertible generative model, conditioning can be transferred to the latent space. Let \(T_0:\R^d\to\R^d\) be an invertible transport map such that \(P=(T_0)_{\#}\pi_0\), where \(\pi_0\) is a simple reference distribution, such as \(\mathcal N(0,I)\). Writing \(x=T_0(z)\) with \(z\sim \pi_0\), the induced posterior distribution on the latent variable \(z\) has density proportional to \(p(y\mid T_0(z))\,\pi_0(z)\). Hence, posterior sampling may be carried out in latent space and then pushed forward through \(T_0\). In the Gaussian-reference case, one may sample the latent posterior by Langevin dynamics
\[
dz_t=-\bigl(z_t+\nabla_z \mathcal L_y(T_0(z_t))\bigr)\,dt+\sqrt{2}\,dW_t,
\]
and then recover posterior samples in data space through \(x_t=T_0(z_t)\). In this way, a generative model trained to realize transport between distributions can also be used to construct posterior distributions under partial observation.

Unlike the conditional-generation setting above, the target distribution here is not a predictive family indexed by context, but a posterior law obtained by updating a prior through an observation model.

\subsection{Mean-field game formulations}

A further extension arises when the distributional shift is generated endogenously by a continuum of interacting agents, as in mean field games (MFGs). MFGs were introduced by \citet{lasry2007mean} and independently by \citet{huang2006large}; for a broad modern treatment, see \citet{carmona2018probabilistic}, and for computational approaches in high-dimensional settings, see \citet{ruthotto2020machine}.

In a first-order deterministic MFG, a candidate population trajectory \(\rho=(\rho_t)_{t\in[0,1]}\) enters each agent's objective, while each agent chooses a velocity field \(\tilde v_t\) and induced distribution trajectory \(\tilde\rho=(\tilde\rho_t)_{t\in[0,1]}\). Starting from an initial distribution \(P\), admissible pairs \((\tilde\rho,\tilde v)\) satisfy the continuity equation \eqref{eq:liouville}, with \(\tilde\rho_0=P\). Given \(\rho\), the representative-agent cost is
\begin{eqnarray}\label{J_def}
&&\mathcal J(\tilde\rho,\tilde v;\rho)\\
&&=
\int_0^1 \int \left(\frac12\|\tilde v_t(x)\|^2 + F[\rho_t](x)\right)\, d\tilde\rho_t(x)\,dt
+ \int G[\rho_1](x)\, d\tilde\rho_1(x),
\nonumber
\end{eqnarray}
where \(F\) and \(G\) denote running and terminal couplings. A best response to \(\rho\) is a minimizer of \(\mathcal J(\tilde\rho,\tilde v;\rho)\) over the admissible class. Intuitively, the representative-agent cost is the expected total cost faced by a single infinitesimal agent when the aggregate population trajectory $\rho=(\rho_t)_{t\in[0,1]}$ is held fixed. The agent controls its own velocity field $\tilde v_t$ and therefore its own induced trajectory $\tilde\rho_t$. The term $\frac12\|\tilde v_t(x)\|^2$ penalizes movement or control effort. The running coupling $F[\rho_t](x)$ captures the cost of being at state $x$ at time $t$ when the rest of the population is distributed according to $\rho_t$; in applications, this may represent congestion, competition, exposure to risk, or interaction effects. The terminal coupling $G[\rho_1](x)$ captures the final cost of ending at state $x$ when the population ends according to $\rho_1$. Thus, the best-response problem asks how one agent would move against a fixed population flow, and the mean-field equilibrium requires this best response to reproduce the same aggregate population flow. A mean-field Nash equilibrium is then a fixed point \((\hat\rho,\hat v)\) such that \((\hat\rho,\hat v)\) is a best response to the population trajectory \(\hat\rho\).

For example, in traffic routing or fleet repositioning, $\rho_t$ may describe the spatial distribution of vehicles over time. A representative vehicle chooses how quickly and where to move, paying effort cost for movement and congestion cost that depends on where other vehicles are located. The resulting deployment distribution is endogenous: it is generated by the collective best responses of many agents rather than imposed externally.

This extends the earlier transport constructions by allowing the objective to depend on the full trajectory, a terminal cost, and interactions through the evolving population distribution. Under special choices of \(F\) and \(G\), closely related formulations reduce to dynamic optimal transport or mean-field control. From an algorithmic viewpoint, one may parameterize the velocity field by a neural ODE or flow-matching network and compute equilibria by iterating between particle evolution in Lagrangian coordinates and updates of the population distribution; see, e.g., \cite{min2021signatured,yu2025high}.

\subsection{A common particle-based implementation scheme}\label{sec:particle}

The distributional problems considered above are often formulated in terms of probability distributions, transport maps, or velocity fields. In practice, however, they can often be approached through a common particle-based scheme. Starting from particles \(\{x_i\}_{i=1}^N\) sampled from an initial distribution \(P\), one approximates the relevant objective by sample averages and solves the resulting finite-dimensional optimization problem over the particle system. Depending on the application, this optimization may represent transport toward a target distribution, adversarial perturbation for robust generation, posterior adjustment under an observation model, or a best-response update in a mean-field game. In each case, the output is a collection of updated particle positions, or more generally, particle trajectories, that represent the optimized distributional shift.

Once these particle trajectories are obtained, one may interpolate them in time and fit a continuous velocity field \(v_t\) by flow matching. In this way, the empirical particle evolution is lifted to an Eulerian description, yielding a continuous-time generative model whose induced distribution flow approximates the optimized particle system. The learned velocity field can then be used to transport new samples generated from the updated distribution, thereby providing a reusable representation of the distributional transformation beyond the original finite particle set.

Viewed in this way, flow matching becomes more than a standalone generative model: it serves as a general implementation mechanism for sample-based optimization in probability space. The optimization is carried out on particles, with objectives estimated via sample averages, while flow matching converts the resulting particle evolution into a continuous transport representation. This provides a common computational template for a broad class of distribution-construction problems under distributional shift.

\paragraph{Toy example: worst-case generation.}
Let \(P=\mathcal N(0,I_2)\) on \(\R^2\), and let \(x_1,\dots,x_N\sim P\). Suppose failure is associated with large values of the first coordinate, so take \(\ell(x)=x^{(1)}\). The robust particle update solves
\[
\max_{x_1',\dots,x_N'}\;\frac1N\sum_{i=1}^N\left[\ell(x_i')-\frac{1}{2\lambda}\|x_i'-x_i\|^2\right].
\]
For this linear loss, the optimizer is explicit: \(x_i'=x_i+\lambda e_1\), where \(e_1=(1,0)\). Thus, the worst-case particle cloud is a shifted version of the nominal one. Interpolating each pair \((x_i,x_i')\) and fitting a velocity field by flow matching yields a continuous transport from the nominal distribution to the worst-case distribution. This illustrates the general scheme: optimize particles by a sample-average objective, then lift the resulting particle system to a continuous-time generative model. This simple construction is also reminiscent of adversarial robustness formulations based on worst-case perturbations of observed samples; see, e.g., \citet{sinha2017certifying}.

\rev{Computationally, the transport-based constructions considered here should be distinguished
from solving exact discrete optimal-transport problems between large empirical measures. In the
implementations emphasized in this tutorial, the main cost is shifted to sample-based optimization
and neural function approximation. For flow matching, training resembles supervised regression
on sampled interpolation points, while sampling requires numerical integration of the learned ODE.
For particle-based robust generation, the main cost comes from alternating updates of the decision
variable and the transported particles, followed, if desired, by fitting a reusable velocity field.
Thus, the cost depends mainly on the number of samples or particles, gradient steps, network size,
and ODE/SDE solver evaluations. Once learned, the resulting transport or velocity field can generate
new scenarios without re-solving the original distributional optimization problem.}

\rev{Practical implementation also involves model-design trade-offs. Flow matching offers a simple
regression-based objective for learning velocity fields; diffusion models can be more stable but may
require more sampling steps; normalizing flows allow likelihood evaluation but impose invertibility
constraints; and particle-based methods are transparent but scale with the number of particles and
inner updates. Key hyperparameters include particle count, network size, learning rate, solver
accuracy, noise or interpolation schedules, and perturbation parameters such as \(\lambda\) or
\(\gamma\). These choices should be tuned using task-level criteria such as held-out decision loss,
stress-test performance, calibration, or constraint violation.}

\rev{Scaling generative models also requires attention to data. In OR applications, useful training data
may come not only from historical observations, but also from simulators, digital twins, controlled
perturbations, or stress-test engines. When data are scarce or high-dimensional, one should exploit
problem structure rather than fit a fully unstructured model. Architectures that share parameters
across time, nodes, scenarios, or contexts can benefit more effectively from large datasets and often
generalize better under shift. In practice, scalable generative modeling is therefore a joint design
problem: collect or simulate data that cover the operational regimes of interest, and choose model
classes whose inductive biases match the structure of the OR system.
}

\section{Theory: Guarantees for distribution construction}
\label{sec:theory}

A useful way to analyze generative procedures for distribution construction is to view them as optimization algorithms in the space of probability distributions. This perspective is especially natural for an OR audience: rather than treating the model as a black-box generator, one interprets each layer, time step, or iterative update as moving a distribution toward a task-dependent objective in probability space. Optimization over distributions has deep geometric roots. Early work on information geometry established a differential-geometric view of families of probability distributions and their associated optimization principles \citep{amari2008information,amari2016information}. A complementary line of work, based on Wasserstein geometry and gradient flows in metric spaces, led to variational formulations of evolution equations and their discrete-time approximations via proximal schemes such as JKO \citep{jordan1998variational,ambrosio2005gradient}. More recently, these ideas have been connected to modern optimization and sampling algorithms on spaces of measures, with applications ranging from variational inference and sampling to generative modeling and distributionally robust optimization \citep{wibisono2018sampling,kent2021modified,cheng2024convergence,xu2024flow}.

The purpose of this section is not to present a single unified theorem for all constructions in Section~5, but to highlight three representative \emph{theoretical templates} for distribution construction. Different tasks specify the target distribution in different ways, which naturally leads to different types of guarantees. For data imitation, the representative result is a contraction-type guarantee in Wasserstein space. For robust generation, it is first-order convergence of a min--max optimization in transport-map space. For posterior updating under partial observation, it is an error-transfer bound from prior approximation and numerical sampling to posterior error. Table~\ref{tab:theory-templates} summarizes these three cases. More broadly, these examples illustrate that different distribution-construction tasks lead naturally to different analytical templates, rather than to a single universal theory.

\begin{table}[t]
\centering
\caption{Representative theory templates for different distribution-construction tasks.}
\label{tab:theory-templates}
\begin{tabular}{p{0.20\linewidth} p{0.28\linewidth} p{0.36\linewidth}}
\toprule
Task & Source of distribution shift & Representative guarantee \\
\midrule
Data imitation &
Transport from data distribution \(P\) toward reference distribution \(Q\) &
Contraction / forward--reverse convergence in Wasserstein space \\

Robust generation &
Adversarial perturbation of a nominal distribution &
First-order convergence of min--max optimization in transport-map space \\

Posterior updating &
Partial observation through a prior--likelihood update &
Error-transfer bounds from prior approximation and sampling error to posterior error \\
\bottomrule
\end{tabular}
\end{table}

\subsection{Iterative flow models as optimization in Wasserstein space}

We begin with the basic data-imitation setting, in which the goal is to generate from a data distribution \(P\) by transporting it toward a simple reference distribution \(Q\), and then reversing the learned transport. Let \(p\) and \(q\) denote the densities of \(P\) and \(Q\), respectively. The key insight is that the forward flow may be viewed as a discretization of a gradient flow in probability space. In regimes where this forward process converges linearly to \(Q\), the number of iterations needed to achieve error \(O(\varepsilon)\) is of order \(\log(1/\varepsilon)\). For iterative flow models, this translates into a logarithmic bound on the required depth \(N\), namely the number of flow steps or residual blocks.

Consider a sequence of invertible maps \(F_1,\dots,F_N\). Here \(F_n\) denotes the \(n\)-th invertible update in the iterative flow model, and the overall transport map is given by the composition of these updates. Starting from the data distribution \(P\), define the forward iterates \(\rho_0=P\) and \(\rho_n=(F_n)_{\#}\rho_{n-1}\), so that
\begin{eqnarray}\label{eq:fwd-bwd-process}
&& \text{(forward)} \qquad
\rho_0=P \xrightarrow{F_1} \rho_1 \xrightarrow{F_2}\cdots \xrightarrow{F_N} \rho_N \approx Q,
\\
&& \text{(reverse)} \qquad
 P\approx \tilde\rho_0 \xleftarrow{F_1^{-1}} \tilde\rho_1 \xleftarrow{F_2^{-1}} \cdots \xleftarrow{F_N^{-1}} \tilde\rho_N = Q.
\end{eqnarray}
Here \(\tilde\rho_n\) denotes the reverse iterates starting from \(\tilde\rho_N=Q\), and \(\tilde\rho_0\) is the generated distribution. If densities exist, we write \(p_n\) and \(\tilde p_n\) for the densities of \(\rho_n\) and \(\tilde\rho_n\). Thus, the analysis separates naturally into two parts: first, show that the forward process drives \(\rho_n\) toward \(Q\); second, transfer this guarantee to the reverse process, which produces the generated distribution \(\tilde\rho_0\).

To analyze this forward process, we compare it with the idealized Wasserstein proximal iteration. We work in Wasserstein space and consider the objective
with the Kullback-Leibler (KL) divergence
\[
G(\rho):=\KL(\rho\|Q) =: \int \rho(x)\log \frac{\rho(x)}{q(x)} dx,
\]
defined for distributions \(\rho\in\mathcal P_2\) that are absolutely continuous with respect to \(Q\). When \(\rho\) admits a density, also denoted by \(\rho\), and \(Q\) has density \(q(x)\propto e^{-V(x)}\), this may be written as
\begin{eqnarray}
    &&G(\rho)=\mathcal H(\rho)+\mathcal E(\rho),\\
&&\mathcal H(\rho)=\int \rho(x)\log \rho(x)\,dx,
\qquad
\mathcal E(\rho)=c+\int V(x)\,d\rho(x),
\end{eqnarray}
for a constant \(c\). This decomposition is useful because the generalized convexity properties of \(G\) can be studied through the entropy term \(\mathcal H\) and the potential-energy term \(\mathcal E\). The Jordan--Kinderlehrer--Otto (JKO) scheme \citep{jordan1998variational} defines an iterative update by
\begin{equation}\label{eq:JKO-obj-1}
\rho_{n+1}\in
\arg\min_{\rho\in\mathcal P_2}
\left\{
G(\rho)+\frac{1}{2\gamma}W_2^2(\rho_n,\rho)
\right\},
\end{equation}
where \(\gamma>0\) is the step size. This is the Wasserstein analog of a proximal step in Euclidean optimization: it balances a decrease in the objective \(G\) against the transport cost of moving away from the current iterate \(\rho_n\).

The convergence of \eqref{eq:JKO-obj-1} depends on the appropriate notion of convexity in Wasserstein space. For the results used here, one requires convexity along generalized geodesics, a standard strengthening of displacement convexity. Many functionals arising in applications satisfy this condition, including entropy, potential energy, interaction energy, and, in particular, the KL divergence under suitable regularity assumptions. Under such conditions, the JKO iteration converges toward the minimizer \(Q\), and under a suitable strong-convexity condition, this convergence is linear. Consequently, after \(N\) iterations,
\[
\KL(\rho_N\|Q)=O(\varepsilon^2)
\qquad\text{for}\qquad
N\asymp \log(1/\varepsilon).
\]
Thus, optimization theory in Wasserstein space yields a depth bound for iterative flow models: logarithmically many steps suffice to drive the forward process close to the reference distribution.

The reverse guarantee follows from invertibility. Because each \(F_n\) is invertible, the reverse process is obtained by pushing \(Q\) backward through the inverse maps, so the forward and reverse terminal distributions are related by bijective transport. A key fact is that \(f\)-divergences are invariant under invertible measurable transformations. Since KL divergence is an \(f\)-divergence, control of the forward error at the terminal time transfers directly to the generated distribution:
\(
\KL(\rho_N\|Q)=\KL(P\|\tilde\rho_0),
\)
or equivalently, at the density level,
\(
\KL(p_N\|q)=\KL(p\|\tilde p_0).
\)
Therefore, the \(O(\varepsilon^2)\) control for the forward process implies the same \(O(\varepsilon^2)\) control for the reverse process. By Pinsker's inequality, this further yields an \(O(\varepsilon)\) bound in total variation. In this way, convergence of the optimization scheme in probability space translates directly into a quantitative density-learning guarantee for the generative model. Thus, the guarantee quantifies how well iterative transport toward a reference distribution approximates the desired data distribution.

On the computational side, related Wasserstein-gradient-flow constructions for generative modeling and inference include sliced Wasserstein flows, large-scale JKO-type methods, and continuous function approximations; see, e.g., \cite{liutkus2019sliced,mokrov2021large,xu2022jko}.
\subsection{Robust generation as optimization in transport-map space}

The previous subsection studied iterative generative models through a minimization problem in the space of distributions, using Wasserstein geometry and the JKO scheme to analyze convergence of the forward and reverse processes. Robust generation leads to a different, but closely related, optimization viewpoint. Here, the problem is no longer a single-objective minimization over distributions, but a min--max problem in which an adversary selects a worst-case distribution. When the nominal distribution \(P\) is absolutely continuous, Brenier's theorem implies that each candidate adversarial distribution \(Q\) can be represented as the pushforward \(Q=T_{\#}P\) of \(P\) under a transport map \(T\). Thus, the inner maximization over distributions is converted into an optimization over transport maps.

Recall that the robust objective is
\begin{eqnarray*}
&&\min_{\theta\in\Theta}\max_{Q\in\mathcal P_2}
\left\{
R_Q(\theta)-\frac{1}{2\lambda}W_2^2(P,Q)
\right\},
\end{eqnarray*}
where \(R_Q(\theta):=\E_{Y \sim Q}[\ell(\theta,Y)]\), and \(\lambda>0\) controls the strength of the adversarial perturbation. Writing \(Q=T_{\#}P\), and equivalently \(Y=T(U)\) with \(U\sim P\), the inner objective becomes
\[
L(\theta,T)
=
\E_{U\sim P}
\left[
\ell(\theta,T(U))
-\frac{1}{2\lambda}\|T(U)-U\|^2
\right].
\]
Unlike the previous subsection, where the iterate is a distribution in Wasserstein space, the present formulation treats the transport map \(T\) as the optimization variable. The Wasserstein penalty again plays the key structural role: it induces the quadratic regularization in \(T(U)-U\) that makes first-order analysis possible.

Assuming that \(\ell(\theta,y)\) is differentiable in both arguments, the gradients of \(L\) with respect to \(\theta\) and \(T\) are
\begin{eqnarray*}
&&\nabla_\theta L(\theta,T)
=
\E_{U\sim P}\,\nabla_\theta \ell(\theta,T(U)), \\
&&\nabla_T L(\theta,T)(u)
=
\nabla_y \ell(\theta,T(u))-\frac{1}{\lambda}(T(u)-u),
\quad P\text{-a.s. } u,
\end{eqnarray*}
where \(\nabla_T L(\theta,T)\) denotes the \(L^2(P)\)-gradient with respect to the transport map \(T\). Starting from an initial pair \((\theta^{(0)},T^{(0)})\), this leads to the gradient descent--ascent iteration
\begin{equation}\label{eq:GDA-transport}
\begin{cases}
\theta^{(k+1)}
=
\theta^{(k)}-\tau\,\nabla_\theta L(\theta^{(k)},T^{(k)}),\\[0.3em]
T^{(k+1)}
=
T^{(k)}+\eta\,\nabla_T L(\theta^{(k)},T^{(k)}).
\end{cases}
\end{equation}
This is the functional analog of gradient descent--ascent in Euclidean minimax optimization \citep{lin2020gradient,yang2022faster}. Under standard smoothness assumptions, and under a nonconvex--strongly-concave or, more generally, nonconvex--PL condition in the \(T\)-variable, one obtains the familiar \(O(\varepsilon^{-2})\) iteration complexity for reaching an \(\varepsilon\)-stationary point.


The transport-map formulation also admits an equivalent intrinsic first-order condition at the level of distributions \cite{xu2024flow}. Fix \(\theta\), and suppose the inner maximization attains a local maximizer \(Q^\star\in\mathcal P_2^r\). Then the corresponding optimality condition in Wasserstein space takes the form 
\[
\nabla_y \ell(\theta,y)+\frac{1}{\lambda}\bigl(T_{Q^\star}^P(y)-y\bigr)=0,
\qquad Q^\star\text{-a.e. } y,
\]
where \(T_{Q^\star}^P\) denotes the optimal transport map from \(Q^\star\) to \(P\). 
If, in addition, the forward optimal transport map \(T^\star:P\to Q^\star\) exists and is the a.e. inverse of $T_{Q^\star}^P$, 
then composing with \(T^\star\) yields
\[
\nabla_y \ell(\theta,T^\star(u))
-\frac{1}{\lambda}\bigl(T^\star(u)-u\bigr)=0,
\qquad P\text{-a.s. } u,
\]
which is exactly the stationarity condition \(\nabla_T L(\theta,T^\star)=0\) in transport-map space. Thus, the ascent direction in \eqref{eq:GDA-transport} is not merely an algorithmic construction; it is the map-space representation of the Wasserstein first-order optimality condition for the worst-case distribution. The above guarantee quantifies how well first-order optimization in transport-map space constructs an adversarially generated distribution.

Although the theory is formulated directly in transport-map space, practical implementations are often particle-based, as discussed in Section \ref{sec:particle}. One samples particles \(U_i\sim P\), approximates the objective and gradients by sample averages, and updates the transported particles \(T^{(k)}(U_i)\). The resulting transported particles can be lifted to a continuous neural network representation by a matching loss, which allows efficient generalization outside the finite training samples.



\subsection{Posterior sampling with generative priors: an illustrative error-transfer guarantee}

Posterior sampling under partial observations leads to a different style of guarantee from the optimization-based results above. In the previous subsections, the main results took the form of contraction or first-order convergence statements for distributional dynamics. Posterior sampling with generative priors has also been studied extensively, particularly using score-based and diffusion models for inverse problems; see, e.g., \citet{song2022solving,chung2023diffusion}. Related theory-oriented analyses are also beginning to appear for posterior-sampling algorithms built on score-based generative priors \citep{sun2024provable}. Here we present one representative example of an \emph{error-transfer guarantee}: approximation error in the generative prior propagates to approximation error in the posterior, together with an additional contribution from numerical sampling.


As one concrete example, let \(P\) denote the true prior of an unknown variable \(U\), and let \(\tilde P=(T_0)_{\#}\pi_0\) be an approximate prior induced by a pre-trained invertible generator \(T_0\) from a simple reference distribution \(\pi_0\). Given an observation \(h\), let \(P^h\) and \(\tilde P^h\) denote the corresponding true and model posteriors obtained by weighting \(P\) and \(\tilde P\) by the likelihood \(p(h\mid U)\). Pulling \(\tilde P^y\) back through \(T_0\) yields a latent posterior on \(z\), which can then be sampled numerically and mapped back to data space through \(U=T_0(z)\). A result in \citep{purohit2024posterior} shows that this construction admits a two-part error decomposition. First, prior approximation error transfers to posterior error: if \(\TV(P,\tilde P)\le \varepsilon\), then \(\TV(P^h,\tilde P^h)\le 2\kappa_h\varepsilon\), where \(\kappa_h\) depends on the likelihood and the true prior. Second, if the numerical sampler approximates the latent posterior with error \(\varepsilon_{\mathrm{samp}}\), and \(\hat P^h\) denotes the resulting sampled posterior in data space, then
\[
\TV(P^h,\hat P^h)\le 2\kappa_h\varepsilon+\varepsilon_{\mathrm{samp}}.
\]
Thus, in this setting, posterior error separates into a prior-modeling term and a sampling term.

\section{OR tasks under distributional shift and examples}
\label{sec:or-tasks}

The earlier sections provided the mathematical and computational tools for constructing probability distributions. We now return to the OR perspective and ask a more operational question: what distribution does the downstream decision problem actually need? Under a distributional shift, the answer is rarely the raw historical distribution. Planning requires a scenario distribution, robustness requires a stress distribution, partial information requires a conditional or posterior distribution, and deployment under regime change requires a corrected distribution.

This section organizes the main OR tasks under that lens. The goal is not to provide an exhaustive survey of applications, but to identify the distributional objects that arise most naturally in OR and to connect them to the constructions introduced earlier. 

\rev{
The two examples below are intended as implementable case studies rather than
large-scale benchmarks. They focus on scenario generation for planning and stress
testing for robust decision-making, rather than on exhaustive empirical comparison.
For each example, we report the data representation, preprocessing, model
architecture, training and sampling procedure, and evaluation criteria. The goal is
to make the distribution-construction pipeline concrete: starting from a nominal
distribution, constructing a decision-relevant distribution, and evaluating whether
the resulting scenarios or decisions improve task-level performance.
}

\subsection{Scenario generation for planning}

In stochastic optimization, the decision problem is often written as \eqref{OR_model} or approximated through a finite scenario set \(\{\xi^i\}_{i=1}^m\) sampled from a distribution \(Q\). The quality of the resulting decision depends not only on the number of scenarios, but also on whether \(Q\) captures the dependence structure, tail behavior, and regime variation relevant to the objective and constraints.

Generative models enter naturally as \emph{scenario engines}. Rather than treating \(Q\) as a fixed historical distribution, one may construct a scenario distribution adapted to the planning context, for example, by conditioning on observed covariates, transporting a nominal distribution to a new regime, or enforcing structural constraints such as temporal, spatial, or network dependence. The point is not merely to match marginals, but to construct scenario distributions under which the downstream optimization behaves reliably.

\paragraph{Example: power outage scenario generation.}
We illustrate scenario generation for planning using county-level power outage data \rev{from \url{PowerOutage.us}} for the Atlanta metropolitan area. \rev{This example is motivated by electric-utility storm preparation and restoration planning. In practice, utilities use outage scenarios to support decisions such as crew staging, mutual-aid requests, equipment allocation, and service-risk assessment. Historical outage data provide a nominal distribution \(P\), while the planning distribution \(Q\) may need to reflect current conditions, regime changes, or stressed but plausible outages. A generative scenario model can support this workflow by producing spatially coherent county-level outage scenarios for downstream resource-allocation and reliability-planning models.} 

Each sample is a 10-dimensional outage-count vector observed at a 15-minute timestamp, whose coordinates correspond to Fulton, DeKalb, Cobb, Gwinnett, Clayton, Douglas, Cherokee, Forsyth, Henry, and Rockdale counties. \rev{Writing the raw vector as \(c_{ij}\in\{0,1,2,\ldots\}\), we set \(y_{ij}=\log(1+c_{ij})\) and \(x_{ij}=(y_{ij}-\mu_{yj})/\sigma_{y,j}\), \(j=1,\ldots,10\), where \(\mu_y\) and \(\sigma_y\) are countywise training means and standard deviations.} The dataset covers 2018--2024 and contains 245{,}472 timestamp-level samples. For expositional simplicity, we treat the timestamped observations as i.i.d. cross-sectional outage vectors and use the example to study spatial dependence across counties rather than temporal dynamics. \rev{Here, spatial dependence refers to the contemporaneous dependence among the ten county-level outage counts observed at the same timestamp, rather than temporal dependence across timestamps. This dependence is incorporated through the joint representation: each training sample is the full vector \(x_i \in \mathbb{R}^{10}\), and a single multivariate flow is trained on these vectors rather than fitting separate one-dimensional models for each county. Thus, the learned velocity field acts on the entire county-level outage vector and can capture cross-county correlations and other joint dependence patterns.} Overall, the example is intended to illustrate multivariate scenario construction under a simplified setting, not to provide a full spatio-temporal outage model. \rev{Since this example focuses on scenario generation rather than a specified downstream planning model, the empirical historical scenario distribution serves as the natural nominal baseline. Accordingly, we evaluate the generated scenarios against the empirical outage distribution using distributional diagnostics, including MMD, marginal empirical CDFs, and cross-county correlation matrices, rather than comparing separate downstream decisions under nominal and shifted deployment distributions.}

We train a continuous-time flow-matching model to \rev{learn the empirical outage-scenario distribution. We use a standard Gaussian reference distribution \(Z\sim N(0,I_{10})\). For each training sample \(x_i\), reference sample \(z_i\), and time \(t\sim \mathrm{Unif}[0,1]\), we use the linear interpolation \(\psi_t=(1-t)x_i+t z_i\) and train the velocity field by minimizing \(\mathbb{E}_{i,z_i,t}[\|v_\theta(\psi_t,t)-(z_i-x_i)\|_2^2]\). At generation time, samples are drawn from the Gaussian reference distribution and the learned ODE is integrated backward from \(t=1\) to \(t=0\) using reverse Euler integration with 100 steps. The velocity field is parameterized by a multilayer perceptron with width 256, depth 4, and a 64-dimensional sinusoidal time embedding. It is trained for 100 epochs with batch size 1024 using AdamW with learning rate \(10^{-3}\) and weight decay \(10^{-5}\). Generated standardized samples \(\tilde x\) are mapped back to count space by \(\tilde y=\mu_y+\sigma_y\odot \tilde x\) and \(\tilde c=\max\{0,\operatorname{round}(\exp(\tilde y)-1)\}\), where the maximum is applied elementwise.}

Evaluation focuses on two features most relevant to scenario generation: the marginal distribution at each county and the dependence structure across counties. \rev{
These diagnostics are example-specific; Section~8 discusses more general evaluation
principles for generative models used in OR decision problems. We compute MMD and correlation diagnostics in the standardized log-count space, while the marginal empirical CDFs are displayed after mapping samples back to count space.} The learned model achieves a Maximum Mean Discrepancy (MMD) value of \(0.0150\) between real and generated samples in the standardized log-count space. \rev{Specifically, for real standardized log-count samples \(a_1,\ldots,a_m\) and generated standardized log-count samples \(b_1,\ldots,b_n\), we use the biased empirical RBF-kernel estimator
\[
\begin{aligned}
\widehat{\operatorname{MMD}}^2
={}&\frac{1}{m^2}\sum_{i=1}^m\sum_{j=1}^m k(a_i,a_j)\\
&{+}\frac{1}{n^2}\sum_{i=1}^n\sum_{j=1}^n k(b_i,b_j)\\
&-\frac{2}{mn}\sum_{i=1}^m\sum_{j=1}^n k(a_i,b_j),
\end{aligned}
\]
and report its square root as the MMD value. Here \(k(a,b)=\exp(-\|a-b\|_2^2/(2h))\), with bandwidth \(h=\operatorname{median}_{i<j}\|a_i-a_j\|_2^2\) computed from real evaluation samples.}

\begin{figure}[t]
    \centering
    \includegraphics[width=\linewidth]{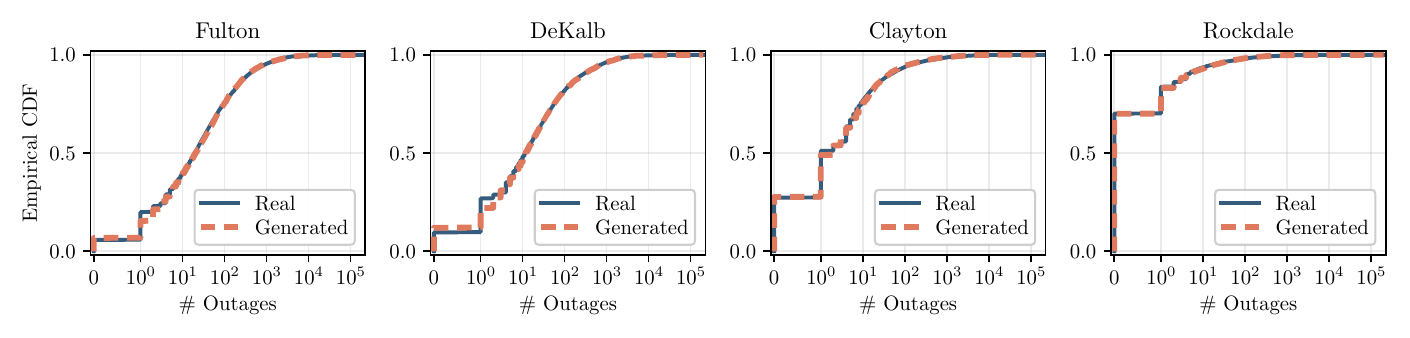}
\caption{Empirical and generated marginal outage distributions for representative counties (Fulton, DeKalb, Clayton, and Rockdale), shown via empirical cumulative distribution functions of outage counts. The four counties illustrate both high-volume and sparse outage regimes.}
    \label{fig:power-marginal}
\end{figure}

\begin{figure}[t]
    \centering
    \includegraphics[width=0.85\linewidth]{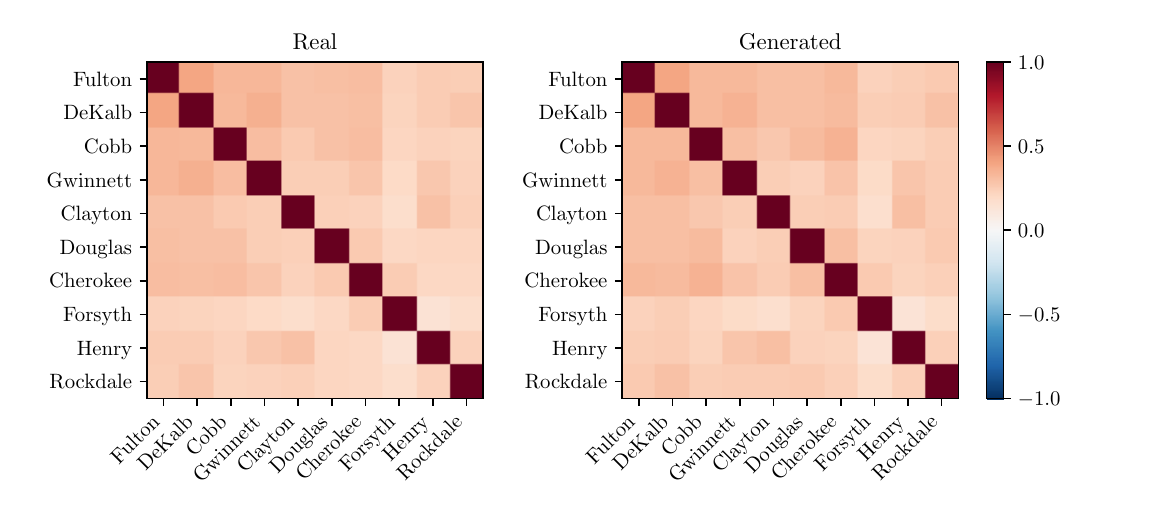}
    \caption{Empirical and generated cross-county correlation matrices in the transformed space.}
    \label{fig:power-corr}
\end{figure}

Figure~\ref{fig:power-marginal} reports representative marginal comparisons for Fulton, DeKalb, Clayton, and Rockdale counties, chosen to span both high-volume and sparse outage regimes. Figure~\ref{fig:power-corr} compares the empirical and generated cross-county correlation structure. The generated samples reproduce the county-wise marginals reasonably well and capture the main pairwise dependence patterns, suggesting that the model can provide realistic multivariate outage scenarios for downstream planning rather than merely matching isolated low-dimensional summaries.

\subsection{Stress testing and distributionally robust decisions}

In robust decision-making, the relevant distribution is not the nominal distribution itself, but a stressed distribution constructed to expose vulnerabilities of the decision system. A standard baseline is Wasserstein distributionally robust optimization, as in \eqref{eq:wass-minimax-0}.

This formulation implicitly characterizes adverse distributions by their membership in an ambiguity set. A constructive alternative is to generate a stressed distribution directly, for example, through transport-guided distortion of \(P\), adversarial updates in transport-map space, or risk-guided dynamics. The result is not only a worst-case value, but an explicit stressed distribution and a corresponding family of scenarios that can be sampled, inspected, and used for downstream analysis. For OR, this matters because robustness is not only about worst-case protection, but also about whether the generated adverse scenarios are operationally meaningful and diagnostically informative.

\paragraph{Example: robust portfolio optimization.}
We illustrate robust distribution construction in a portfolio-allocation problem. The uncertain input is a one-period return vector \(Y\in\R^d\), and the decision is a long-only portfolio \(w(\theta)=\operatorname{softmax}(\theta)\in\Delta^{d-1}\). We use six exchange-traded funds, \(\{\texttt{SPY},\texttt{IWM},\texttt{EFA},\texttt{EEM},\texttt{AGG},\texttt{GLD}\}\), representing U.S. equities, small-cap equities, developed international equities, emerging markets, bonds, and gold. \rev{Using adjusted daily prices from Yahoo Finance,} we construct aligned daily returns and split them chronologically into a training period and a held-out test period. \rev{The data cover 2023--2025, with observations through 2024 used for training. For a raw daily return \(r_i\in\R^6\), let \(\mu\) be the training mean and let \(s=(nd)^{-1}\sum_{i=1}^n\sum_{j=1}^d |R^{\mathrm{train}}_{ij}|\) be a single global scale, where \(n\) is the number of training observations and \(d=6\). The flow model and robust optimization problem are both written in the scaled space \(x_i=(r_i-\mu)/s\). This global scaling keeps the portfolio loss in a common return geometry while avoiding asset-wise rescaling of the decision objective.}

\rev{As in the outage example, the nominal return distribution is learned by flow matching using the same MLP architecture, optimizer settings, batch size, and 100-step reverse Euler sampler. We use a standard Gaussian reference distribution \(\varepsilon\sim N(0,I_6)\), the linear interpolation \(\psi_t=(1-t)x_i+t\varepsilon_i\) between each scaled training return \(x_i\) and a reference sample \(\varepsilon_i\), and the target velocity \(\varepsilon_i-x_i\) in the usual squared flow-matching loss. The flow is trained for 20{,}000 optimization steps. After training, we draw \(N\) i.i.d. nominal return scenarios \(\hat x_1,\ldots,\hat x_N\) from the learned flow, with \(N\) chosen to match the number of training dates.}

The nominal portfolio is obtained by minimizing the scenario average of a smooth shortfall loss. \rev{In the scaled space, define \(\bar q(w)=(q-\mu^\top w)/s\) and \(\ell(x,w)=\beta^{-1}\log(1+\exp(\beta(\bar q(w)-x^\top w)))\), where \(q\) is the target one-period return level. The generated scenarios \(\hat x_i\) are samples from a learned approximation of the empirical nominal distribution. The nominal portfolio, therefore, serves as the no-shift baseline: it optimizes over the learned nominal scenario distribution without adversarial perturbation. The robust portfolios are optimized over adversarially perturbed particles initialized from the same nominal scenarios, so the comparison below evaluates the effect of replacing the nominal no-shift baseline by stress-aware deployment distributions. The nominal problem solves \(\min_\theta N^{-1}\sum_{i=1}^N \ell(\hat x_i,w(\theta))\). For robust optimization, we use the particle form
\[
\min_\theta \max_{z_1,\ldots,z_N}
\frac{1}{N}\sum_{i=1}^N
\left[
\ell(z_i,w(\theta))-\frac{1}{2\gamma}\|z_i-\hat x_i\|_2^2
\right],
\]
where \(z_i\) are adversarially perturbed return particles in the same scaled space. We use \(q=0.02\), \(\beta=4\), and \(\gamma\in\{0.01,0.1,1,10\}\). We initialize \(\theta=0\), corresponding to the equal-weight portfolio, and initialize \(z_i=\hat x_i\). The nominal portfolio is optimized for 100{,}000 gradient steps, and each robust portfolio is optimized for 100{,}000 gradient descent--ascent steps. The softmax parameterization enforces the long-only simplex constraint, so no projection is needed for \(w\).}

\begin{figure}[t]
    \centering
    \includegraphics[width=0.8\linewidth]{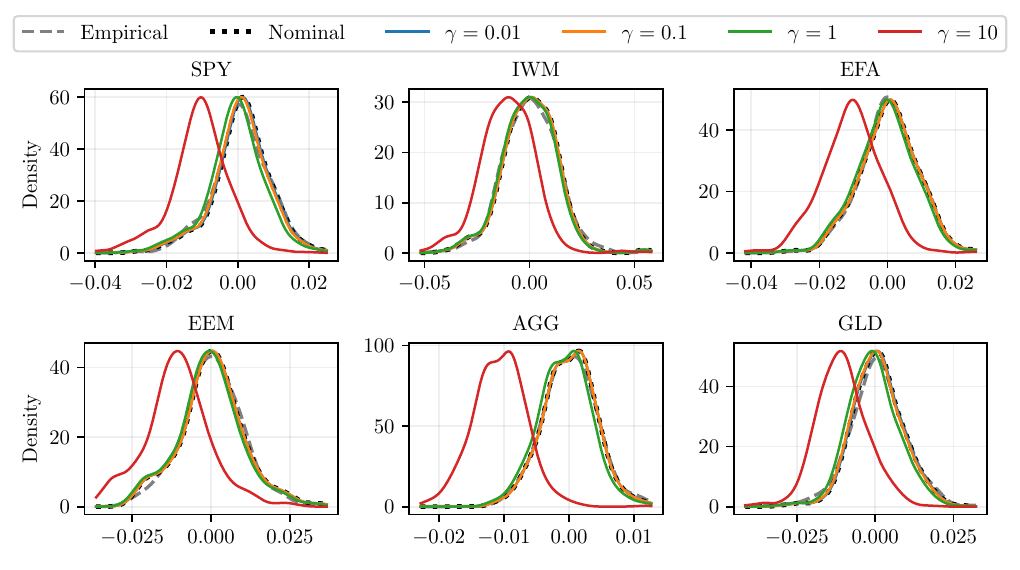}
    \caption{Marginal distributions of nominal and generated worst-case returns for the six assets, shown for different values of the regularization parameter. Larger perturbations produce more pronounced shifts toward adverse return regions.}
    \label{fig:stress-placeholder-1}
\end{figure}

\begin{figure}[t]
    \centering
    \includegraphics[width=0.7\linewidth]{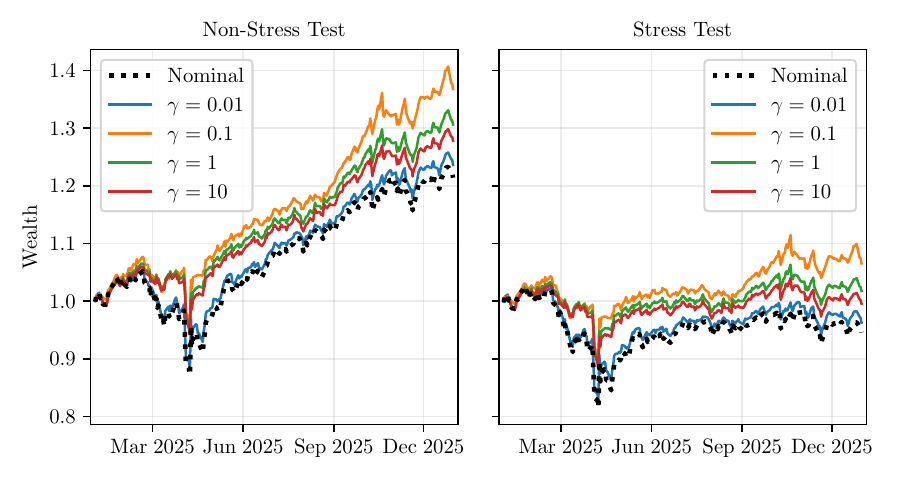}
    \caption{Out-of-sample cumulative wealth of the nominal portfolio and portfolios optimized under generated worst-case distributions, evaluated on a non-stress test setting (left) and a stress test setting (right). Curves correspond to different values of the regularization parameter. These examples illustrate that $\gamma = 0.1$ achieves the best performance among the choices.}
    \label{fig:stress-placeholder-2}
\end{figure}

Figure~\ref{fig:stress-placeholder-1} shows the generated worst-case return distributions for different values of the regularization parameter \(\gamma\). \rev{Larger values of \(\gamma\) weaken the quadratic penalty and allow larger perturbations away from the nominal scenarios.} As the perturbation becomes less constrained, the stressed distribution shifts away from the nominal one and places more mass on adverse return regions. \rev{Figure~\ref{fig:stress-placeholder-2} compares out-of-sample cumulative wealth on the held-out test set for this baseline and for portfolios optimized under the generated worst-case distributions. The comparison includes the nominal portfolio, which optimizes over generated nominal scenarios without adversarial perturbation, and robust portfolios optimized over the generated worst-case particles. We evaluate each portfolio under the original held-out test returns and under a simple market-wide stress test obtained by shifting each held-out return vector downward by \(10^{-3}\mathbf{1}\). This stress transformation is intentionally simple and is used as a stress-level diagnostic rather than a comprehensive heterogeneous market-stress model.} Overall, the example illustrates how generated adverse distributions can be used to construct stress-aware decisions in a concrete OR setting, rather than merely reproducing historical returns.

\rev{The main computational costs are transparent in this particle representation. Sampling \(N\) scenarios with \(K\) reverse Euler steps requires \(K\) velocity-field evaluations per scenario, while each full-batch robust update changes both the portfolio parameter and the \(N\) adversarial particles. Since the present examples are low-dimensional, they are intended to clarify the pipeline rather than benchmark large-scale training.}

\subsection{Conditional and posterior updating under partial information}

Many OR problems are sequential: information arrives over time, and the relevant uncertainty distribution must be updated before each decision. In this setting, the decision-maker typically needs either a conditional distribution indexed by observed side information, or a posterior distribution obtained through a prior--likelihood update after observing a noisy signal. These two objects serve different roles: conditional distributions describe predictive uncertainty indexed by context, whereas posterior distributions arise from Bayesian updating under partial observation.

This task is central in monitoring, filtering, anomaly detection, and sequential intervention, where decisions depend on calibrated updates rather than unconditional sample generation. The relevant uncertainty is no longer the unconditional historical distribution, but an updated distribution tailored to the information available at decision time.

\subsection{Transport-based correction across regimes}

A final task is correction across regimes. Here, the issue is not stress or partial observation, but deployment mismatch: the historical distribution used for training differs systematically from the distribution under which the decision will be deployed. Examples include market regime shifts, geographic transfer, policy changes, altered sensing pipelines, or new operational environments.

In such settings, the central question is how to construct a corrected distribution \(Q\) from a nominal distribution \(P\). When the mismatch is well described as a geometric deformation, a natural representation is \(Q=T_{\#}P\), where \(T\) transports the nominal distribution to the deployment regime while preserving structural features that should remain stable. When the shift is largely explained by observed covariates, the same problem may instead be formulated through conditional distributions indexed by regime variables. From an OR standpoint, this matters because optimization under an uncorrected historical distribution may produce systematically biased or poorly calibrated decisions at deployment.

These last two tasks further emphasize the central theme of this section: the key question is not only how to construct a distribution, but which distribution the decision problem actually requires.

Accompanying code for reproducing the numerical examples in this section are available at \url{https://github.com/xieyao14/informs-tutorial26/}, including data preprocessing, training, sampling, and evaluation scripts.

\section{Evaluation}
\label{sec:evaluation}

A central message of this tutorial is that, in OR, a generative model should be evaluated by the \emph{distribution it constructs for the decision problem}, not by generic sample quality alone. Once generative modeling is viewed as distribution construction rather than unconstrained synthesis, evaluation becomes inherently task-dependent.

Evaluation should be determined by the distributional operation of interest and the downstream decision criterion. One should first determine what distribution the decision problem requires, then assess whether the chosen representation constructs that distribution well enough for the task at hand.

For scenario generation in stochastic optimization, the relevant question is whether the constructed distribution captures the dependence structure, tail behavior, and regime variation that materially affect the decision. Appropriate diagnostics, therefore, include feasibility under held-out scenarios, service levels, reserve violations, stockouts, and out-of-sample cost, rather than only visual realism or low-dimensional goodness of fit. For stress testing and robust decision-making, the target is different: the constructed distribution should reveal adverse yet plausible failure modes. Evaluation should then focus on tail loss, constraint violation under stress, decision sensitivity, and comparison with baseline adversaries such as Wasserstein DRO.

For conditional generation and posterior updating, the constructed distribution should be calibrated relative to the available information. Relevant diagnostics include conditional coverage, posterior uncertainty quantification, and the stability of updates as new observations arrive. In sequential settings, these translate into operational criteria such as false alarm rates, detection delay, intervention timing, and downstream utility. For transport-based correction across regimes, the key question is whether the corrected deployment distribution improves decision quality under shift; useful diagnostics include held-out deployment performance, robustness across regimes, and calibration under regime change.

Across these settings, the most informative criterion is often \emph{decision regret}: the loss incurred by optimizing under the constructed distribution rather than under the true deployment distribution. Exact regret is rarely observable, but it can often be estimated through held-out deployment periods, common stress scenarios, or controlled perturbation experiments. This is usually the most direct test of whether distribution construction has improved the downstream decision. By contrast, generic metrics from data-imitation tasks, such as likelihood or perceptual sample quality, are at best secondary: they may be useful for some representations, but they do not by themselves determine whether the constructed distribution is decision-relevant.

\section{Conclusion and discussion}
\label{sec:conclusion}

A main message of this tutorial is that generative models in OR should not be viewed solely as tools for imitating historical data or as black-box sample generators. Their broader value lies in providing flexible mechanisms for constructing probability distributions under structured distributional shift. This perspective is especially relevant in decision-making under uncertainty, where the distribution needed by the downstream problem is often not the nominal historical distribution itself, but a derived distribution obtained through transport, conditioning, guidance, or equilibrium effects.

From this viewpoint, generative modeling becomes part of the modeling language of OR. It provides tools for constructing scenario distributions for planning, stress distributions for robustness, updated distributions under partial information, and corrected distributions under regime change. The emphasis, therefore, shifts from data imitation alone to the broader problem of distribution construction: how to represent, compute, and validate the uncertainty distribution that is actually relevant to the decision at hand.

This tutorial also highlights that these constructions admit meaningful mathematical structure. In several settings, generative procedures can be interpreted through optimization in probability space, whether through Wasserstein-space minimization, min--max optimization in transport-map space, or posterior error transfer under Bayesian updating. At the same time, the strongest current guarantees remain largely at the level of idealized distributions, transport maps, and particle systems, while the gap to finite-sample, stochastic, and neural implementations remains substantial.

This leaves substantial room for future OR research. On the theoretical side, sharper foundations are needed for structured distribution construction, especially in settings involving partial observation, sequential updating, constraints, and interacting agents. On the algorithmic side, generative models offer a promising way to enrich modern OR methods, not only by producing samples but by enabling new optimization, control, and decision procedures in high-dimensional and data-rich settings. More broadly, generative modeling points toward a view of uncertainty that is constructive, task-aware, and better aligned with the needs of modern decision-making in OR.

\rev{Finally, this perspective should not be read as a recommendation to use generative models in every OR problem. Generative models are not always necessary. In low-dimensional, data-limited, or
well-structured problems, empirical distributions, parametric models, kernel density
estimation, or bootstrap methods may be simpler, more transparent, and equally
effective. Such simpler methods are preferable when the decision depends mainly on low-dimensional
summaries such as means, variances, or quantiles. Generative models are most useful
when the decision-relevant distribution involves complex dependence, conditioning,
tail behavior, high-dimensional structure, or structured distributional shift, and when the training samples are abundant. In
practice, they should be used only when they improve task-level metrics such as
decision loss, calibration, robustness, or constraint violation relative to simpler
baselines.
}

\section*{Acknowledgement}
The work of Y.Z. and Y.X. was partially supported by an NSF CMMI-2112533, and the Coca-Cola Foundation. Y.X. and X.C. were partially supported by NSF DMS-2134037. XC was also partially supported by NSF DMS-2237842, DMS-2031849, and the Simons Foundation MPS-MODL-00814643.

\bibliographystyle{informs2014}
\bibliography{references,flow}

\newpage

\end{document}